\documentclass[PHME, 2024]{PHMSociety}

\usepackage{graphicx}
\usepackage{amsmath}
\usepackage{amssymb}
\usepackage{lipsum}
\usepackage{multirow}
\usepackage{multicol}
\usepackage{makecell}
\usepackage{subfigure}
\usepackage[table]{xcolor}
\usepackage{float}
\restylefloat{table}

\begin{document}

\title{Continuous Test-time Domain Adaptation for Efficient Fault Detection under Evolving Operating Conditions}

\author{%
	Han Sun\authorNumber{1}, Kevin Ammann\authorNumber{2}, Stylianos Giannoulakis\authorNumber{3}, and Olga Fink\authorNumber{4}
}

\address{
	\affiliation{{1,4}}{EPFL, Ecublens, Vaud, 1024, Switzerland}{ 
		{\email{han.sun@epfl.ch}}\\ 
		{\email{olga.fink@epfl.ch}}
		} 
	\tabularnewline 
	\affiliation{2, 3}{Sulzer, Winterthur, Zürich, 8401, Switzerland}{ 
            {\email{Kevin.Ammann@sulzer.com}} \\
		{\email{stylianos.giannoulakis@sulzer.com}}
		} 
}

\maketitle
\pagestyle{fancy}
\thispagestyle{plain}

\phmLicenseFootnote{Han Sun}

\begin{abstract}
Fault detection is crucial in industrial systems to prevent failures and optimize performance by distinguishing abnormal  from normal operating conditions. Data-driven methods have been gaining popularity for fault detection tasks as the amount of condition monitoring data from complex industrial systems increases. Despite these advances,  early fault detection remains a challenge under real-world scenarios. The high variability of operating conditions and environments makes it difficult to collect comprehensive training datasets that can represent all possible operating conditions, especially in the early stages of system operation. Furthermore, these variations often evolve over time,  potentially leading to entirely new data distributions in the future that were previously unseen. These challenges prevent direct knowledge transfer across different units and over time, leading to the distribution gap between training and testing data and inducing performance degradation of those methods in real-world scenarios.
To overcome this, our work introduces a novel approach for continuous test-time domain adaptation. This enables early-stage robust anomaly detection by addressing domain shifts and limited data representativeness issues. We propose a Test-time domain Adaptation Anomaly Detection (TAAD) framework that separates input variables into system parameters and measurements, employing two domain adaptation modules to independently adapt to each input category. This method allows for effective adaptation to evolving operating conditions and is particularly beneficial in systems with scarce data. Our approach, tested on a real-world pump monitoring dataset, shows significant improvements over existing domain adaptation methods in fault detection, demonstrating enhanced accuracy and reliability.
\end{abstract}

\section{Introduction}
\label{introduction}

Fault detection aims to identify evolving faults or degradation in complex industrial
systems, aiming to prevent system failures or malfunctions. Early and robust fault detection is essential for optimizing equipment performance and minimizing maintenance and unavailability costs.
Recently, data-driven methods have been widely applied to fault detection facilitated by the growing availability of system condition monitoring data \cite{fink2020potential}. 
However, these methods often assume the availability of abundant, representative training datasets to learn a data distribution that is applicable across all relevant operating and environmental conditions. Such representative training datasets are frequently not available due to the high diversity of systems and the wide range of operating conditions. This issue is particularly acute for newly installed or refurbished units with limited observation periods. 
A potential solution to this problem is to transfer knowledge and operational experience from fleet units with extensive and relevant data to those lacking representative training data. This approach leverages the rich experience and datasets of 'experienced' units to enhance the learning and performance of less experienced ones, aiming to bridge the gap in data availability and representativeness across the fleet. However, such knowledge transfer might lead to insufficient performance, as data-driven methods typically assume identical and independent distributions (i.i.d) between training and testing data, which does not hold true in real-world industrial complex systems with varying operating conditions and dynamic environments. This leads to significant discrepancies in data distribution between fleet units. Consequently, a model trained on one unit may perform poorly when applied to another, evidenced by a high rate of false alarms, preventing them from benefiting from the existing fleet knowledge.

A substantial amount of research has been performed to address such a challenge by applying domain adaptation (DA) approaches \cite{yan2024comprehensive}, which aim to bridge the domain shift between a labeled source and a related unlabeled target domain. However, The scarcity of faulty data in industrial systems introduces specific challenges for DA in fault detection because industrial applications typically lack labeled source data for supervised learning.  Furthermore, these methods usually assume discrete source and target domains.
However, the operating conditions of complex systems evolve over time, leading to continuous domain shifts within the same unit. Therefore, domain adaptation should not only occur between units but also be continuously applied within a unit, rather than assuming a single, discrete target domain, to ensure robust fault detection. 

In this work, we propose a novel approach for fleet-wide continual test-time domain adaptation, aiming to achieve robust anomaly detection across different units of a fleet over time. 
Our proposed fault detection framework, based on signal reconstruction, integrates a domain adaptive module specifically designed to address the dynamic and evolving environments of complex industrial systems. This approach aims to enhance robustness and adaptability in fault detection within these challenging contexts.
To prevent overfitting to the faulty data distribution during adaptation, we categorize the input variables into two groups: control parameters and sensor measurements. We then integrate two domain adaptive modules to adapt to the data distribution of each category separately. This strategy enables us to distinguish between normal variations inherent within the systems and abnormal changes in operating status, thereby improving the accuracy of our anomaly detection framework.
By integrating adaptation into the basic fault detection pipeline, TAAD facilitates the transfer of operational experience between different units of a fleet, thereby benefiting from the collective knowledge of the fleet. 
TAAD has been evaluated on a real-world pump monitoring dataset, and the results demonstrate notable improvements compared to other domain adaptation methods. Our proposed framework is transferable to other industrial applications and enables more timely and robust fault detection in complex industrial systems.






\section{Related Work}

\subsection{Fault Detection in Prognostics and Health Management} \label{fault detection}
Prognostics and Health Management (PHM)  seeks to enhance equipment performance and minimize costs by enabling 
precise detection, diagnosis, and prediction of the remaining useful lifetime as accurately as possible. It integrates the detection of an incipient fault (fault detection), its isolation, the identification of its origin, and the specific fault type (fault diagnostics), along with the prediction of the remaining useful life \cite{fink2020potential}. Fault detection aims to identify faulty system conditions based on current operating conditions and gathered condition monitoring data. The complexity of real-world industrial systems poses specific challenges for achieving accurate and robust fault detection. First, faulty data is scarce in real industrial systems. Failures in critical systems, such as power or railway systems, are infrequent. Furthermore, it often takes a considerable amount of time for a system to degrade to the point of failure or the end of life.
As a result,  faults are often never or seldom encountered during limited time periods and are therefore absent in training datasets.

Consequently, one of the main research directions in fault detection has focused on unsupervised learning, which can be categorized into three main directions\cite{ruff2021unifying}. 
Probabilistic models aim to approximate the normal data probability distribution. The estimated distribution mapping function can then be used as an anomaly score. Different deep statistical models have been applied for probability-based anomaly detection, such as energy-based models (EBMs) \cite{zhai2016deep}.
One-class classification models directly learn a discriminative decision boundary that corresponds to a desired density level of normal samples, instead of estimating the full density \cite{ruff2021unifying}. This approach aims to learn a compact boundary that encloses the normal data distribution \cite{wang2018unsupervised, zhang2021anomaly}.
Reconstruction-based methods learn a model, such as autoencoders (AEs), are optimized to reconstruct the normal data samples well and detect via reconstruction error \cite{lai2023context, hu2022low}. These models are expected to fit the data distribution under healthy conditions and then raise an alarm for predictions with large deviations when the test data distribution is significantly different from the learned distribution. Thus, the reconstruction error serves as an anomaly score for detecting faults.

Other studies have focused on semi-supervised learning, where it is presumed that a limited number of faulty data samples are accessible for training \cite{ramirez2023semi}.

\subsection{Fleet Approaches for Fault Detection} \label{fleet}
Unsupervised fault detection, as discussed in section \ref{fault detection}, relies on the assumption that all possible normal conditions of the system can be learned from a sufficiently large and representative training dataset. 
However, collecting a dataset representative enough for new systems or refurbished units to cover all possible normal operating conditions within a short time period is unlikely. While extending the observation period can facilitate the collection of more comprehensive data, it also hinders early monitoring of the system.
In such cases, transferring operational experience from other similar units with longer and more representative data can significantly enhance robust detection at an early stage. These units can be grouped into a fleet, where each unit shares similar characteristics \cite{leone2017data}. A good example would be a fleet of gas turbines or cars produced by the same manufacturer, albeit with different system configurations, operating under varying conditions in different parts of the world \cite{fink2020potential}.

The direct transfer of fleet knowledge assumes  identical and independent distributions (i.i.d) between
training and testing units.
However, this assumption often does not hold for complex industrial systems, which are characterized by varying operating conditions and changing environments. This discrepancy poses a significant challenge in transferring a developed model across different units within the fleet. Traditional methods focus on identifying units that are similar enough to form sub-fleets \cite{leone2016data, liu2018ai, michau2018fleet, michau2019unsupervised}. Such methods depend on the entire fleet sharing sufficient similarity and fail when units under homogeneous conditions do not exist or cannot be identified. Recently, domain adaptation has been used to transfer knowledge between units or between different operating conditions within the same unit \cite{yan2024comprehensive}, a topic discussed in section \ref{domain adaptation}.

\subsection{Domain Adaptation Applied to Fault Detection} \label{domain adaptation}
A substantial amount of research in the field of PHM has focused on domain adaptation, a subtopic of transfer learning, including discrepancy-based methods \cite{zhang2022class, qian2023maximum} and adversarial-based methods \cite{michau2021unsupervised, qian2023deep, nejjar2024domain}. 
These DA methods aim to align the data distribution between the source and target domains, assuming that the target samples available are abundant enough to represent target data distribution. However, this assumption does not hold true for newly installed systems with limited data samples collected, which prevents prompt system monitoring as discussed above \cite{michau2021unsupervised}.
Furthermore, these methods typically assume one or more discrete, static target domains and attempt to adapt to them. However, operating conditions often evolve continuously over time, potentially leading to unseen distribution shifts in the future. Therefore, it is necessary to continuously adapt to domain shifts on the fly, rather than assuming a single discrete target domain \cite{wang2022continual}.

Test-time adaptation (TTA) aims to adapt a source-pretrained model to a target domain without using any source data. The model is dynamically updated on the fly, based on the current data batch, without exposure to the entire target data set. 
Representative methods utilize batch normalization, estimating and normalizing mean and variance on each batch to update the model \cite{wang2021tent, liang2020we}. 
Thus, TTA can be applied to adapt batch data online, accommodating continuous domain shifts for fault diagnosis \cite{wang2019domain}. 
Although this branch of methods can be directly adapted for the fault diagnostic task, it is not suitable for unsupervised fault detection. 
In scenarios of unsupervised fault detection, where detection is based on deviation from the norm, applying TTA to the current batch of data with unknown labels may cause the model to unintentionally fit potentially faulty data within this batch. Consequently, the anomalies might not be recognized as out-of-distribution, leading to a reduced ability of the model to identify faults based on prediction errors. 

\subsection*{}

To conclude, robust fault detection in PHM at the early stage encounters the challenge of data scarcity. Fleet approaches help units that are newly taken into operation benefit from fleet knowledge, while their transferability is constrained by the high variability of system operating conditions within the fleet. 
Current DA methods applied for fault detection cannot simultaneously address all of the challenges we discussed above. They either fail to adapt to continuous domain shifts or are incompatible with the limited data and label availability elaborated above.




\section{Methodology}
\label{methodology}

\subsection{Problem Definition}

The primary motivation of this research is to transfer knowledge from one system, which has abundant monitoring data, to other systems or fleets operating under varying conditions. 
Often, these systems and fleets are newly taken into operation, for which only a limited amount of observations can be collected for training. Their data distribution can evolve continuously due to changes in operating conditions and environmental factors. The objective is to adapt the prediction model trained on the original system, enabling it to make accurate predictions for new systems and fleets, even when only a few training samples are available. Given:
\begin{itemize}
    \item abundant healthy training data from the source system: $X_s = \left[x_1^s, \cdots, x_n^s\right]$, where $s$ denotes the source domain and $n$ denotes the number of data samples from the source domain, and
    \item limited observed normal data from the target domain: $X_t = \left[x_1^t, \cdots, x_m^t\right]$, where $t$ denotes the target domain and $m$ denotes the number of available data samples from the target domain,
\end{itemize}
the goal here is to achieve robust fault detection in the target domain $t$.

The proposed method takes into account limited data availability and varying operating conditions, specifically addressing scenarios where: 1) no anomalies are available for training; 2) only limited target data is available for adaptation; and 3) continuous changes in operating conditions occur during test time. 

\subsection{Reconstruction-based Anomaly Detection Framework} \label{reconstruction framework}
\begin{figure}[ht]
    \centering
    \includegraphics[width=0.95\linewidth]{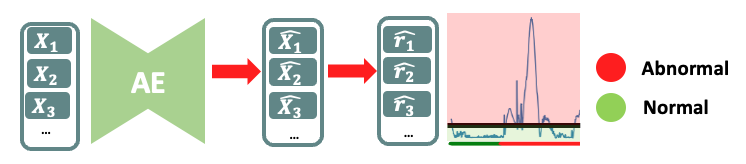}
    \caption{General pipeline of reconstruction-based unsupervised anomaly detection}
    \label{fig:pipeline}
\end{figure}

We develop a reconstruction-based anomaly detection pipeline, which achieves robust fault detection by continuously adapting to novel operating conditions, as depicted in Figure \ref{fig:pipeline}. 
This approach utilizes an autoencoder (AE), denoted as $f_\theta$, trained exclusively on normal source data samples, $X_s$, for the purpose of signal reconstruction. The goal of $f_\theta$ is to accurately model the normal data distribution of $X_s$ with accurate predicted signal value $\hat{X_s}$. The training objective is to minimize the mean-squared error (MSE) between the original data samples $X_s$, and their reconstructed counterparts  $\hat{X_s}$:
\begin{equation}
    loss_{MSE}=\frac{1}{n} \sum_{1}^{n}\left(X_s-\hat{X}_s\right)^2
\end{equation}
where $n$ denotes the number of training samples.
Thus, on the healthy source dataset, we expect a small residual value $\hat{r_s} = \hat{X_s} - X_s$.
During testing, data samples generating large residuals are considered out-of-distribution and subsequently labeled as anomalies.

The autoencoder architecture consists of two parts: an encoder $f_e$ and a decoder $f_d$. $f_e$ comprises three fully-connected layers each followed by a batch normalization layer and a ReLU activation function, which consecutively map the original signal input to feature dimensions of 50, 50, and 10. $f_d$ follows a similar architecture but without batch normalization layers, decoding the latent representation from 10 to 50, 50, and then back to the original signal dimension. 

\subsection{Anomaly Score and Anomaly Detection} \label{anomaly detection}
During test time, we compute the fault label $y\in[0, 1]$ based on the reconstruction result. $0$ denotes a healthy sample while $1$ indicates a faulty sample. Given the $i_{th}$ data sample $X_i = [x_i^1, .... x_i^k]$, we compute its relative residual:

\begin{equation}
\centering
    r_i = \frac{|\hat{X_i} - X_i|}{\bar{X}_{t\_training}}
\end{equation}

given its predicted reconstruction result $\hat{X_i}$. $k$ indicates the input dimension. $\bar{X}_{t\_training}$ represents the mean value of target data samples for training (including validation data), which helps scale the residual values. The anomaly score $s_i$ is calculated by integrating the scaled residual values across all sensors:
\begin{equation}
    s_i = \frac{1}{k} \sum_{j=1}^{k}r_i^j + max\sum_{j=1}^{k}r_i^j
\end{equation}

To avoid false detection by outliers with extremely large residuals, the computed anomaly score is smoothed within a certain window length $l$:
\begin{equation}
    s_{i\_smooth} = min\sum_{q=0}^{l-1}s_{i+q}
\end{equation}

Anomalies are then detected based on $s_{i\_smooth}$, using a threshold determined via statistical analysis of the healthy validation set. We identify the data sample $X_i$ as an anomaly if:
\begin{equation}
    s_{i\_smooth} > \alpha * \bar{r}_{t\_training}
\end{equation}
where $\alpha$ is set empirically with a trade-off between the reduction of false alarms and sensitivity to faults.

In this case study, potential faults are reported and examined daily. Thus, the evaluation of abnormal conditions is conducted on a daily basis, where we compute the number of cumulative abnormal data samples of the day.

\subsection{System Variables} \label{system variables}
Directly applying domain adaptation to the current anomaly detection framework can potentially cause the model to fit unknown abnormal samples in the target domain's current batch during test time, thus impairing the model's ability to detect those faults. To distinguish between data distribution shifts due to changing operating conditions and occurrences of abnormal operating status, we split the input parameters into two groups: $X = [x, w]$.
$w$ denotes control variables, indicating variables that control system conditions. These variables are set by the operators or by the control system to optimize the performance under specified conditions.
$x$ represents sensor measurements, which are sensor signals monitoring system components and reflecting real-time system states. 
Here, we assume that changes in the distribution of control variables do not necessarily indicate an abnormal status but rather distinct operating conditions to which we should adapt.

\subsection{Test-time Domain Adaptation Anomaly Detection}
\begin{figure*}[t]
    \centering
    \includegraphics[width=0.90\linewidth]{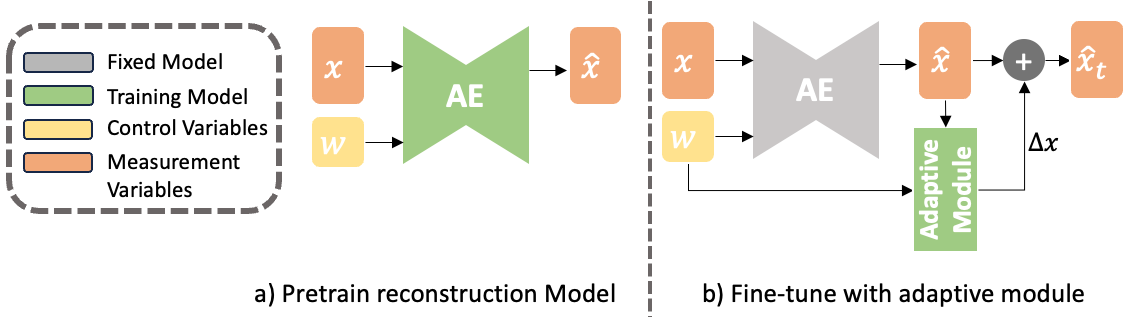}
    \caption{Test-time domain Adaptation Anomaly Detection (TAAD). a) We first pretrain the reconstruction-based anomaly detection model on the source dataset; b) For domain adaptative prediction, we add domain adaptive module and train it on the target training data.}
    \label{fig:framework}
\end{figure*}

Figure \ref{fig:framework} illustrates our proposed cross-domain, reconstruction-based anomaly detection framework TAAD, inspired by recent advancements in test-time domain adaptation. This framework enables us to achieve robust anomaly detection across different domains through online adaptation. 
Based on the pretrained reconstruction framework introduced in \ref{reconstruction framework}, we integrate an adaptive module, $h_\phi$, for test-time domain adaptation to bridge the domain gap between the source and target domains. 
The decision to incorporate a separate adaptive module, $h_\phi$, rather than embedding adaptive layers directly into the reconstruction model, stems from limitations observed in unsupervised anomaly detection. 
TTA methods, such as AdaBN, adapt to each batch during test time, inevitably fitting the distribution of abnormal data points.  This impairs the model's ability to distinguish between normal and abnormal samples. Instead, our adaptive module takes the predicted value and original controlled system variables as inputs. By excluding monitoring data signals from adaptation, this approach prevents overfitting to the potentially faulty data distribution, thereby preserving the model's capability to accurately distinguish anomalies.


The adaptive module is a simple network composed of two fully connected layers that map the group of control variables from its original feature dimension to 10, and then back to its original dimension;  the first is followed by a batch normalization layer and ReLU activation, while the second is followed by ReLU activation only. This adaptive module exclusively processes the control variables $w$ as its input.
During the adaptation phase, the pre-trained autoencoder $f_\theta$ is frozen, and the adaptive module $h_\phi$ is trained on a few target data samples to predict $\Delta x$, aimed at compensating for the large prediction errors due to the domain gap between source and target data. To address continuous domain shifts during test time, an AdaBN layer is incorporated into the adaptive module.  This layer updates its mean and variance based on batch statistics during test time. The predicted $\delta x$ is then added to the original prediction made by $f_\theta$ to compensate for inaccurate predictions caused by operating condition domain shift.

\section{Case Study on Real-world Pump Dataset}
\label{experiments}

\subsection{Industrial Pump Dataset}
In this case study, we aim to achieve early and robust fault detection while reducing false alarms under normal operating conditions. We evaluate our proposed method on a case study of a real industrial dataset,  highlighting its effectiveness in achieving early fault detection and minimizing false positive alarm rates. The experiments are conducted on a real-world pump dataset, which comprises condition monitoring data collected from users with installations of various types of pumps in different locations. As a result, these data represent real-world, noisy data distributions and encompass a wide range of diverse domains, including operating conditions, environments, and pump types. In this dataset, we apply the proposed methodology to obtain robust adaptation across pumps, stations, and different pump types.

The selected dataset has two installation stations, with four Heat Transfer Fluid pumps installed at each, and the pumps are equipped with dual seals of two different types.
Several seal failures were recorded for seven out of eight pumps during the data collection period. This dataset is marked by continuous changes in operating conditions as the controlled parameters are adapted by the operators regularly.  We chose five pumps from this dataset with enough recorded data samples for validation for the case study, as summarized in Table~\ref{tab: sulzer_dataset}. 
To comply with data policy requirements, we use virtual dates (year.month.day) to represent timelines. 
Fault durations reported by on-site operators may lack precision due to delayed inspections. Additionally, faults often occur significantly earlier than actual failures, and systems do not immediately return to normal conditions after maintenance of those failures. Given these factors, we consider data samples within a two-month window of any reported faults as uncertain regarding their health status.  Therefore, we exclude them from both training data and the subsequent evaluation of false alarms.

\begin{table}
\resizebox{0.5\textwidth}{!}{%
\begin{tabular}{ |c||c|c|c|  }
    \hline
    Station & Pump & Seal Type & Operator Maintenance Reports\\
    \hline
    \multirow{2}{1em}A & A-A & Type 1 &  \makecell{primary\& secondary seal replacement, \\primary\& secondary seal leakage} \\
    \cline{2-4}
    & A-C & Type 2 & \makecell{secondary seal replacement, \\primary \& secondary seal leakage} \\
    \hline
    \multirow{3}{1em}B 
    & B-B & Type 1 & \makecell{seal replacement} \\
    \cline{2-4}
    & B-C & Type 2 & \makecell{secondary seal replacement, \\primary  seal leakage} \\
    \cline{2-4}
    & B-D & Type 1 & \makecell{N/A} \\
    \hline
\end{tabular}}
\caption{Details on industrial pump dataset.}
\label{tab: sulzer_dataset}
\end{table}

We categorize the input variables according to the sub-components associated with the pumps of this dataset. 
The studied pumps are composed of four main parts: pump bearings, pump driver axial bearings, motor bearings, and seals.
Since only seal faults are reported in this dataset, we specifically focus on performance and seal-related variables, while disregarding other parameter groups. The used parameters are displayed in table \ref{tab: sulzer_dataset_variables}. 
Sensor monitoring data are available for both the drive end (DE) side and non-drive end (NDE) side of the pumps, reflecting the state of the dual seal on either side and its primary and secondary components.
We exclude pump seal level variables due to their frequent manual adjustments, which do not accurately reflect the operating status of the pump seal. Instead, we focus on variables within the pump performance group, which represent the general operating condition of the current pump. In all our experiments, we designate the seal DE and NDE variables as $x$ and the pump performance group variables as $w$ inputs, as introduced in Section \ref{system variables}.

\begin{table}[ht]
\resizebox{0.5\textwidth}{!}{%
\begin{tabular}{ |c||c|c|}
    \hline
    Group & Description & Variable  \\
    \hline
    \multirow{6}{*}{Pump DE Seal} & \multirow{6}{*}{\makecell{Dual Seal \\ Parameters}} 
    & pump DE seal pressure \\
    \cline{3-3}
    & & pump DE seal pressure secondary \\
    \cline{3-3}
    & & pump DE seal temperature \\
    \cline{3-3}
    & & pump DE seal temperature secondary \\
    \cline{3-3}
    & & \cellcolor[HTML]{C0C0C0}pump DE seal level \\
    \cline{3-3}
    & & \cellcolor[HTML]{C0C0C0}pump DE seal level secondary \\
    \hline
    
    \multirow{6}{*}{Pump NDE Seal} & \multirow{6}{*}{\makecell{Dual Seal \\ Parameters}} 
    & pump NDE seal pressure \\
    \cline{3-3}
    & & pump NDE seal pressure secondary \\
    \cline{3-3}
    & & pump\ NDE seal temperature \\
    \cline{3-3}
    & & pump NDE seal temperature secondary \\
    \cline{3-3}
    & & \cellcolor[HTML]{C0C0C0}pump NDE seal level \\
    \cline{3-3}
    & & \cellcolor[HTML]{C0C0C0}pump NDE seal level secondary \\
    \hline

    \multirow{6}{*}{\makecell{Pump \\ Performance}} & \multirow{6}{*}{\makecell{Pump \\ Operating \\ Conditions}}  & pump uncorrected flow \\
    \cline{3-3}
    & & pump speed \\
    \cline{3-3}
    & & pump pressure suction \\
        \cline{3-3}
    & & pump pressure case \\
        \cline{3-3}
    & & pump uncorrected head \\
        \cline{3-3}
    & & pump uncorrected shaft power \\
    \hline
\end{tabular}}
\caption{Groups of input variables of the industrial pump dataset}
\label{tab: sulzer_dataset_variables}
\end{table}

\subsection{Details on Data Selection and Implementation}
Robust fault detection is critical for newly established industrial systems, posing a challenge due to the brief operational history of such pumps.  Our goal is to enable early fault detection capabilities with minimal data. To address this, we adopt a strategy where a source model is pretrained on a well-established pump with abundant data samples, then adapted to target pumps with limited operational data. This approach aims to achieve robust fault detection on target pumps despite the limited available data for these new installations.

We select pump B-C as our source domain, ensuring that abundant normal data samples are available for training. We train the source model on this pump using data samples from a 12-months period and validate it with an additional 12 months of healthy data.
The remaining pumps are treated as target domains, each with limited training samples. For each pump, we use three months of normal data samples for training and one month for validation. The training and testing data for the target domains are summarized in Table \ref{tab: da_experiment}. 
Due to sensor failures, some data samples lack measurements related to seal components. Considering that only four parameters are available for each seal, with each being crucial and independent of the other parameters, we exclude any data samples with missing measurements. We use the Min-Max Scaler to scale the signals to a range between 0 and 1.

During test time, we combine the 3-month training and 1-month validation data in the target domain to compute the threshold $\bar{r}_{t\_training}$ for anomaly detection. We set $\alpha= 1.5$ for domain adaptation within the same installation station and $\alpha = 2.0$ for adaptation across stations, as the domain gap is comparatively larger.

\subsection{Evaluation Metrics for Robust and Early Detection}
Given the real-world nature of our dataset, which features imbalanced data,  limited collected samples, and uncertain labels for validation,  traditional evaluation metrics such as F1 score and accuracy may not be applicable.
To align with needs and interests in real industrial application scenarios, we evaluate TAAD from two perspectives: 1) minimizing false alarms on normal data samples caused by domain shift and 2) achieving early detection of significant system faults.

Unsupervised fault detection relies on the assumption that the model learns the healthy data distribution and identifies deviating distributions as faults. As the operating conditions of complex industrial systems vary, novel operating conditions that have not been seen by the model are often identified as faults. Such false positive (FP) predictions need to be avoided.
To assess the effectiveness of TAAD in reducing false alarms due to domain shift, we test TAAD on data collected under unseen healthy conditions with only positive samples. Thus, any reported faults are FP.
We identify periods of known normal operations and evaluate the prediction results for these periods, excluding data samples from two months before and after any reported fault to avoid periods of potential pre-fault conditions. The evaluation periods vary for each pump due to data sample availability limitations, as detailed in Table \ref{tab: da_experiment}.
We determine the count of FP, which indicates inaccurately predicted faults, and compute the false positive rate: $\frac{FP}{FP + TP}$ (TP indicates true positives).
The lower the rate, the better our adaption to novel operating conditions and our ability to avoid false alarms.

Faulty conditions vary in severity levels. The system or a specific component can continue to operate despite the occurrence of faults, gradually degrading until a complete failure. This gradual degradation can lead to more severe faults, ultimately stopping operation and potentially leading to secondary damages. Therefore, our goal is to achieve early detection before the faults are observed and recorded.
For early detection of system faults, we summarize all reported faults for each pump in Table \ref{tab: da_experiment}. We conduct an evaluation starting 14 days before the recorded fault date to determine the earliest point of detection achievable by TAAD. This evaluation, performed daily as stated in section \ref{anomaly detection}, focuses on the first predicted abnormal day. This value indicates how early we can detect potential faults in the system and preemptively address them. Additionally, we report the number of days detected as anomalies within this 14-day window to assess detection robustness.  A higher count of abnormal days following the initial fault observation indicates better robust detection capability in consistently issuing alarms, enhancing certainty to involve on-site inspection and minimizing missed faults.


\section{Experimental Results Pump Case Study}
Experiments on fault detection in this pump system, which includes two installation stations, involve two distinct case studies: intra-station transfer, where the domain gap is relatively smaller, and inter-station transfer, which has a larger domain gap.
The proposed method is compared with AdaBN and MMD, as well as with the baseline model without adaptation on the target data.  Performances are reported based on the evaluation metrics introduced in section \ ref{}. General experiment results are summarized in Table \ref{tab: da_experiment_result}.

     
    
    




\begin{table*}[ht]
\centering
\resizebox{0.7\textwidth}{!}{%
\begin{tabular}{ |c|c|c|c|c| }
    \hline
     target domain & \makecell{training time period \\ on target domain} & normal test time period & fault type & faulty time period \\
     \hline
     
    
    B-B & 00.01.01-05.01 & 00.05.01-06.01 & secondary seal replacement & 00.10.07-10.14 \\
    \hline
    
     \multirow{2}{*}{B-D} & \multirow{2}{*}{00.01.01-04.01} & \multirow{2}{*}{00.11.01-01.01.01}
     & secondary seal leakage & 00.08.07-08.22 \\
     \cline{4-5}
     & & & primary seal leakage & 01.11.12-11.13 \\
     \hline

    \multirow{2}{*}{A-A} & \multirow{2}{*}{00.01.01-00.05.01} & \multirow{2}{*}{00.06.01-00.07.01}
    & secondary seal leakage & 00.08.13-08.15 \\
    \cline{4-5}
    & & & primary seal leakage & 00.10.14-11.17 \\
    \hline


    \multirow{2}{*}{A-C} & \multirow{2}{*}{00.01.01-00.05.01} & \multirow{2}{*}{01.07.01-01.08.01}
    & primary seal leakage & 00.07.10-08.10 \\
     \cline{4-5}
    & & & secondary seal leakage & 01.12.26\\
    \hline
\end{tabular}}
\caption{Experimental settings on pump dataset. We report the training time period for domain adaptation for each pump and the normal test time period with no observed faults in between for evaluating the false alarm rate. We report the fault type and the observed faulty time period of each occurrence of fault to evaluate the detection days in advance and the number of detected abnormal days within 14 days prior to the first reported date of the fault. The dates are represented by virtual dates with the same duration and intervals as in the real dataset due to the data privacy policy.}
\label{tab: da_experiment}
\end{table*}

\begin{table*}[ht]
\centering
\resizebox{1.0\textwidth}{!}{%
\begin{tabular}{ |c|c|c|c|c|c|c|c|c|c|c|c|c|c|c| }
    \hline
     \multirow{2}{*}{target domain} & \multirow{2}{*}{\makecell{num of \\ normal samples}} & \multicolumn{4}{|c|}{false alarm rate $\downarrow$} & \multirow{2}{*}{fault type}  &  \multicolumn{4}{|c|}{detection days in advance $\uparrow$} & \multicolumn{4}{|c|}{\makecell{num of detected abnormal days \\ within 14 days $\uparrow$}} \\
     \cline{3-6} \cline{8-15}
     & & baseline & AdaBN & MMD & TAAD & & baseline & AdaBN & MMD & TAAD & baseline & AdaBN & MMD & TAAD  \\
     \hline
     
    
    B-B & 1903 
    & 0.15 & 0.03 & 0.05 & \textbf{0.02}
    & secondary seal replacement & \textbf{14} & \textbf{14} & \textbf{14} & \textbf{14} & \textbf{12} & \textbf{12} & \textbf{12} & \textbf{12} \\
    \hline
    
     \multirow{2}{*}{B-D} & \multirow{2}{*}{2275} & \multirow{2}{*}{\textbf{0}} & \multirow{2}{*}{\textbf{0}} & \multirow{2}{*}{0.01} & \multirow{2}{*}{\textbf{0}}
     & secondary seal leakage & 2 & 0 & 2 & \textbf{4} 
     & 1 & 0 & 1 & \textbf{3}  \\
     \cline{7-15}
     & & & & & & primary seal leakage & 0 & 0 & 0 & \textbf{9} & 0 & 0 & 0 & \textbf{8}\\
     \hline

    \multirow{2}{*}{A-A} & \multirow{2}{*}{1969} & \multirow{2}{*}{\textbf{0}} & \multirow{2}{*}{\textbf{0}} & \multirow{2}{*}{0.01} & \multirow{2}{*}{0.01} 
    & secondary seal leakage & 1 & 0 & 10 & \textbf{11} 
    & 1 & 0 & 2 & \textbf{4}   \\
    \cline{7-15}
    & & & & & & primary seal leakage & 0 & 0 & 0 & \textbf{1}
    & 0 & 0 & 0 & \textbf{1} \\
    \hline


    \multirow{2}{*}{A-C} & \multirow{2}{*}{2135} & \multirow{2}{*}{\textbf{0}} & \multirow{2}{*}{\textbf{0}} & \multirow{2}{*}{0} & \multirow{2}{*}{0.07}
    & primary seal leakage & 0 & 0 & 0 & \textbf{1} 
    & 0 & 0 & 0 & \textbf{2}   \\
    \cline{7-15}
    & & & & & & secondary seal leakage & 0 & 0 & 0 & \textbf{14} 
    & 0 & 0 & 0 & {\textbf{6}} \\
    \hline
\end{tabular}}
\caption{Experimental results on pump dataset. We compare our method with the baseline model w/o domain adaptation, AdaBN, and MMD, and report the false alarm rate and detection days in advance on 4 pumps across two stations. The best results are in bold.}
\label{tab: da_experiment_result}
\end{table*}



\subsection{Case 1: Transfer within Station}

In the first case study, we evaluate the adaptation performance of our TAAD applied to two pumps characterized by a relatively small domain gap. These two pumps are installed at the same station as pump B-C, on which we train the source model.

\begin{figure*}
    \centering
    \subfigure[]{\includegraphics[width=0.80\textwidth]{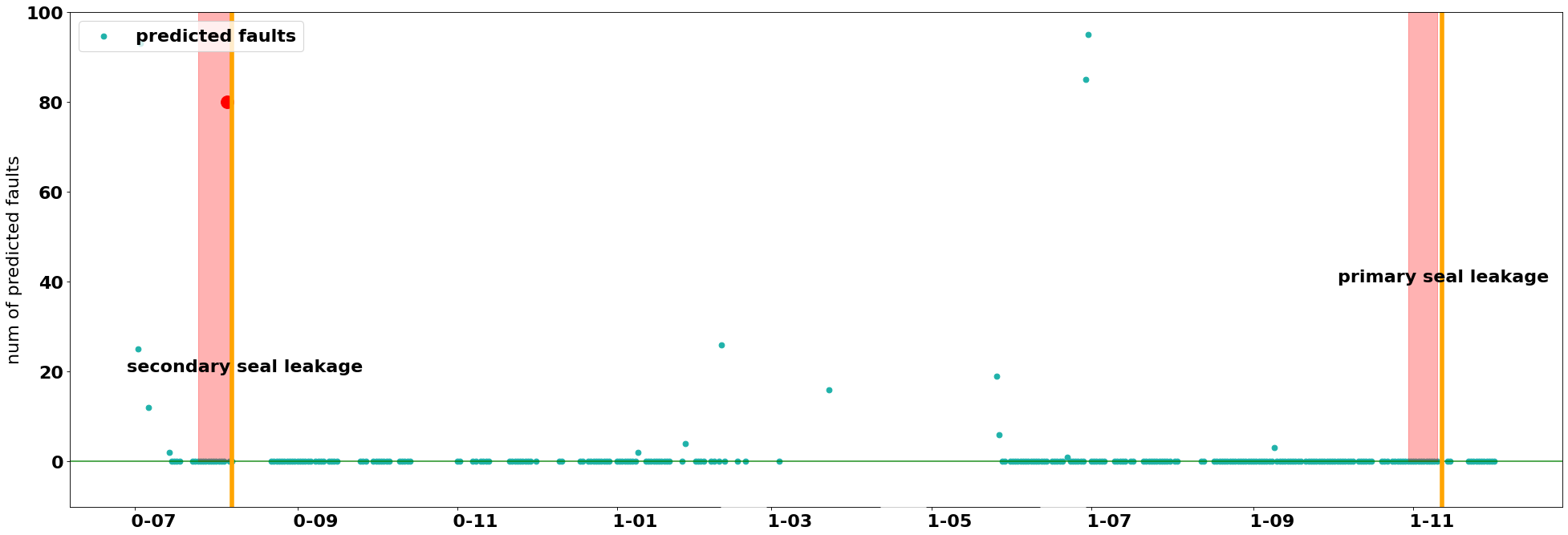}} 
    \subfigure[]{\includegraphics[width=0.80\textwidth]{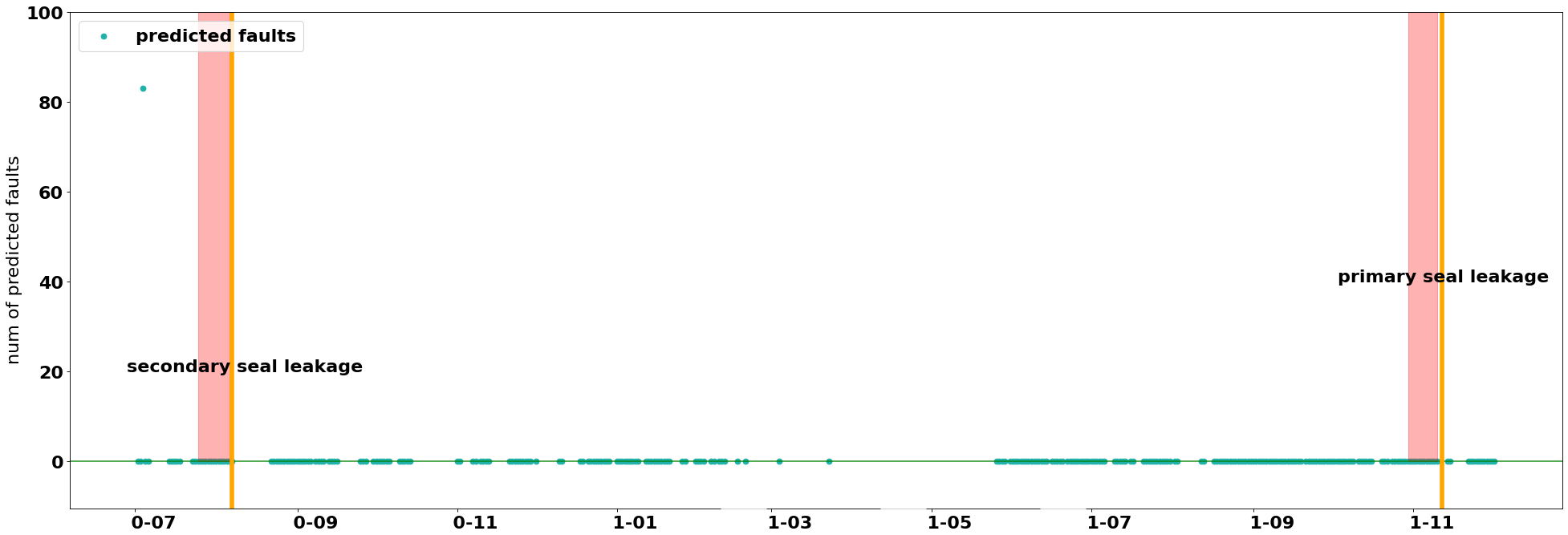}} 
    \subfigure[]{\includegraphics[width=0.80\textwidth]{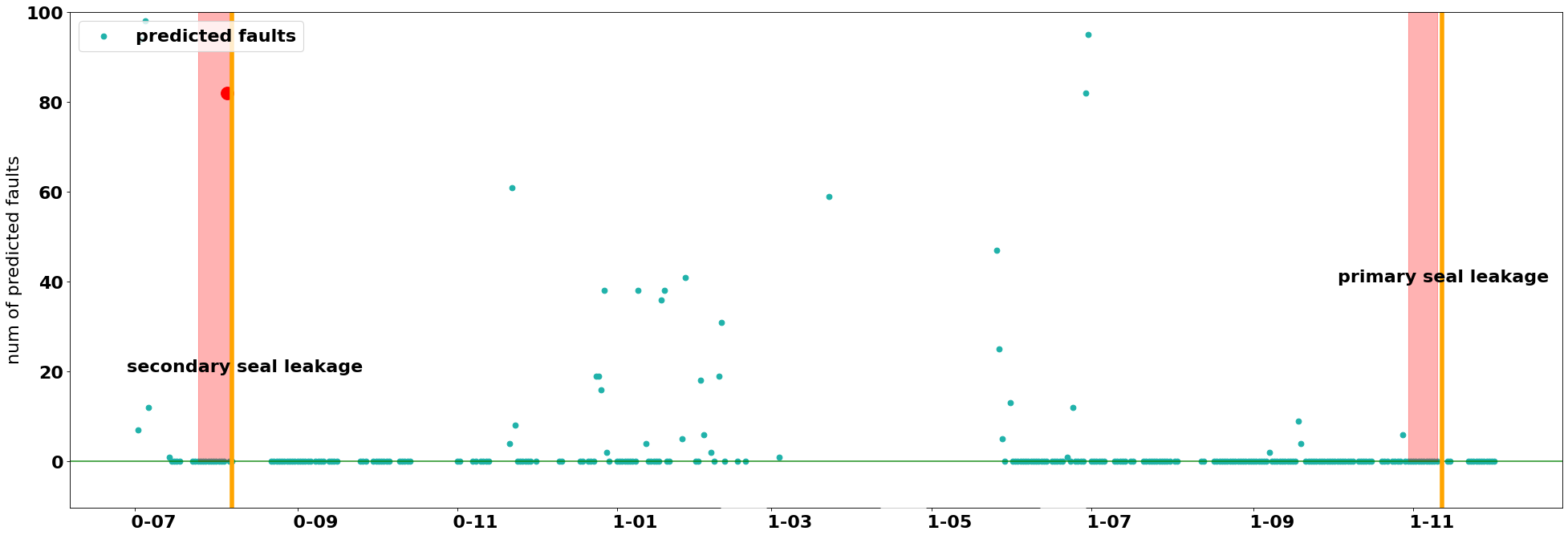}}
    \subfigure[]{\includegraphics[width=0.80\textwidth]{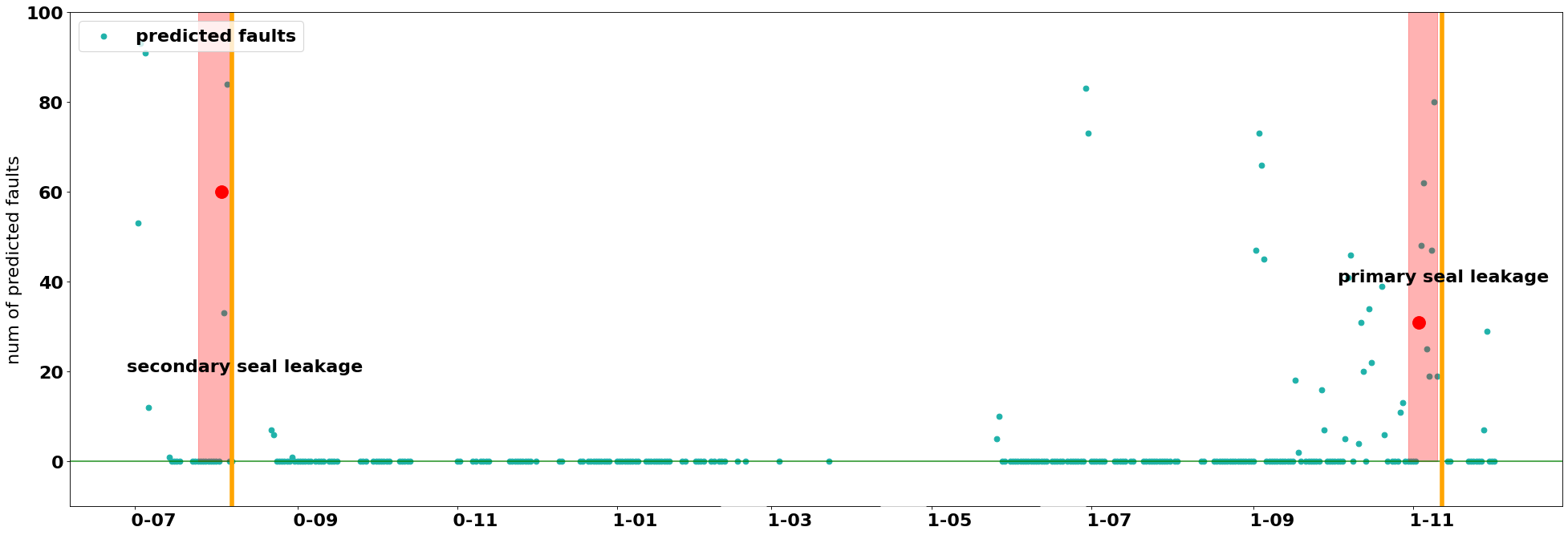}}
    \caption{Comparison of performance of  (a) baseline (b) AdaBN (c) MMD  (d) TAAD for fault detection on Pump B-D. The number of predicted faults per day is plotted. The yellow lines mark the starting date of the reported faults. Red regions mark the 14-day time window before the occurrence of reported faults, and the red dots within this region mark the first detected abnormal day within this period.}
    \label{fig:foobar}
\end{figure*}

\textbf{Pump B-D:} Two seal leakages were observed after the training time period of this target pump. We visualize this case in Figure~\ref{fig:foobar} for a better understanding. The yellow vertical lines mark the recorded first occurrence data of the faults. We scatter the predicted faults of each method on a daily basis. In the case of the earlier secondary seal leakage, occurring  4 months after the adaptation,  all methods managed to detect it, however, TAAD not only detected it but also did so earlier and the detection is more robust with more true positives.  Regarding the later primary seal leakage, which occurred 1 year 7 months after adaptation, all other methods failed to trigger any alarms as the operating condition evolved. In contrast, TAAD successfully detected the reported fault 9 days in advance,  registering  8  abnormal days, thus demonstrating its superior adaptability to long-term changes in operating conditions. As AdaBN does not consider source data and thus tends to overfit current batch statistics, it fails to detect any fault. Furthermore,  compared to MMD, TAAD effectively reduced false alarms, highlighting the robustness of our proposed approach,  which benefits from avoiding overfitting to unstable and unrepresentative measurement variables during adaptation.

\textbf{Pump B-B:} A leakage in the secondary seal is reported 5 months after the training period for adaptation. Given the proximity of the event and the small domain gap, all methods were capable of predicting the fault 14 days in advance. In this scenario, TAAD significantly reduced the false alarm rate from 0.15 to 0.02 by adapting to changing operating conditions. Even when compared to MMD -- which benefits from access to the source domain's training data for direct data distribution alignment -- TAAD demonstrated superior adaptation capabilities by avoiding overfitting to the system's atypical operating status.

\subsection{Case 2: Transfer across Stations}

This case study involves transferring a model pre-trained on pump B-C from station B to station A, anticipating a significantly larger domain gap due to variations in environments and operational regimes across the stations. 

\textbf{Pump A-A:} Two seal leakage faults were reported successively 3 months after the training time period for adaptation on this target pump. For the first secondary seal leakage fault, TAAD achieved the earliest detection with the highest number of days detected as anomalies within this 14-day window before the recorded fault and only 1\% false alarms during the normal operating period. Detecting the subsequent primary seal leakage proved much more difficult, as it occurred shortly after the maintenance of the previous fault and operated under unstable and significantly different operating conditions. Here, while all the compared methods failed to detect the fault, our proposed method managed to raise an alarm on the last day before the fault was reported.

\textbf{Pump A-C:} Two seal leakage faults were recorded. The other methods failed to detect either fault, whereas TAAD successfully reported both. However, in this challenging scenario,  TAAD increased the false alarm rate by 7\% due to the less accurate adaptation.

Generally,  experimental results confirm that detecting faults is more challenging than in intra-station cases. Nonetheless, TAAD successfully detects faults under these challenging scenarios, outperforming other methods.

\subsection{Discussion}
We demonstrate the effectiveness of TAAD to achieve early and robust fault detections in the above two case studies. First, TAAD significantly reduces the false alarm rate under easy-to-detect scenarios compared to other methods, as proved in case 1 on pump B-B, when the target pump is installed within the same station with a smaller domain gap and the fault happens shortly after the adaptation training.
Second, TAAD remains effective and robust long after the initial adaptation phase, as shown in the inboard seal leakage of pump B-D and the outboard seal leakage of pump A-C. Those cases demonstrate the ability of the proposed method to adapt to dynamic evolving operating conditions
Third, TAAD achieves robust detection under significant domain shifts across different installation stations compared to other methods which fail to detect before the occurrence of faults, as shown in the experiments on pumps A-A and A-C.

In general, our TAAD achieves overall better performance than the other methods achieving earlier detections, and providing more continuous and robust detection within the time window before fault occurrences, all while maintaining a low false alarm rate.

\section{Conclusions}

In this paper, we propose an effective continuous test-time domain adaptation approach TAAD for efficient and robust anomaly detection under evolving operating conditions. This approach does not require labeled faulty data and needs only a minimal amount of normal data samples for adaptation. Such requirements align well with the practical needs of real-world industrial systems. We compared our method with two other representative domain adaptation methods. The experimental results demonstrate TAAD's effectiveness in achieving early fault detection under significant domain shifts, both across different stations and over time, while maintaining a low false alarm rate.

Despite its satisfying performance, we see potential improvements in the current method. First, our adaptive module continuously adapts to the current batch without considering the size of the domain gap. We hypothesize that the performance could be enhanced by re-training this module once a significant domain shift is detected. Second, the thresholding parameter $\alpha$ is currently determined empirically. An automatic adjustment of this parameter, taking into account both distribution shifts and operational requirements, could optimize the trade-off between minimizing false alarms and attaining prompt fault detection.

\bibliographystyle{apacite}
\PHMbibliography{PHME_Latex_Template}

\end{document}